\documentclass[10pt,twocolumn,letterpaper]{article}

\usepackage{cvpr}              

\usepackage{colortbl}
\usepackage[table]{xcolor}
\usepackage{multirow}
\definecolor{cvprblue}{rgb}{0.21,0.49,0.74}
\definecolor{oursblue}{rgb}{0.88, 0.93, 0.98}
\usepackage[pagebackref,breaklinks,colorlinks,allcolors=cvprblue]{hyperref}

\def\paperID{****} 
\def\confName{CVPR}
\def\confYear{2026}

\title{UST-Hand: An Uncertainty-aware Spatiotemporal Point Cloud Interaction Network for 3D Self-supervised Hand Pose Estimation}

\author{Tianhao Han$^{1}$, \; Haoyang Zhang$^{5,6}$, \; Liang Xie$^{5,6}$, \; Haochen Chang$^{3}$, \; Kun Gao$^{4}$, \; Yuan Cheng$^{1}$, \\ \; Pengfei Ren$^{2}$\thanks{Corresponding Author.}, \; Erwei Yin$^{1,5,6}$\footnotemark[1]\\
$^{1}$School of Computer Science, Shanghai Jiao Tong University,  \\ $^{2}$Beijing University of Posts and Telecommunications,  \quad
$^{3}$Sun Yat-sen University, \\
$^{4}$Peking University,  \quad  $^{5}$Defense Innovation Institute, Academy of Military Sciences\\
$^{6}$Tianjin Artificial Intelligence Innovation Center\\
{\tt\small tianhao\_jiao@sjtu.edu.cn; rpf@bupt.edu.cn; yinerwei1985@gmail.com}
}

\begin{document}
\maketitle
\begin{abstract}

Manually annotating accurate 3D hand poses is extremely time-consuming and labor-intensive. Existing self-supervised hand pose estimation methods leverage the discrepancy between input images and rendered outputs, or multi-view consistency constraints, as the driving force to optimize networks and progressively refine pose accuracy. However, these methods are highly susceptible to noisy pseudo-labels and overlook the importance of fully exploiting fine-grained spatial correlations, which undermines the stability of model training. To address these issues, we propose UST-Hand, a self-supervised learning framework that estimates uncertainty distribution of hand pose and constructs a probabilistic point cloud feature space, which enables the complex spatiotemporal relationship modeling. UST-Hand employs a conditional normalizing flow model to capture hand pose distributions and samples diverse hypotheses, facilitating robust learning under noisy pseudo-labels supervision with enhanced stability. These multi-hypothesis are mapped to a unified probabilistic 3D point cloud space for multi-view and temporal feature interaction, comprehensively exploring hand motion patterns and fine-grained spatial correlations. Extensive experiments on three challenging datasets demonstrate that UST-Hand achieves state-of-the-art performance, outperforming existing self-supervised methods by up to 37.8\% in Mean Per Vertex Position Error (MPVPE).
\end{abstract}
\section{Introduction}
\label{sec:intro}
Reconstructing 3D hands from images is a fundamental task in numerous human-computer interaction applications such as virtual and augmented reality (AR/VR), action and gesture recognition~\cite{chang2025hierarchical}, and robotic manipulation~\cite{bai2020user, christen2022d, taheri2022goal}. In recent years, various depth-based~\cite{ren2019srn,huang2020awr,ren2021pose,ren2023two} and RGB-based methods~\cite{potamias2025wilor, pavlakos2024reconstructing,ren2023decoupled,zheng2021sar,sun2023smr, ren2025prior, fan2025pose,chen20253} have been proposed for 3D hand pose estimation. However, the advancement of these methods is primarily driven by the availability of large-scale datasets with accurate 3D annotations, which are usually prohibitively expensive and labor-intensive to acquire. This annotation bottleneck significantly limits the scalability of models and hinders their deployment in diverse real-world scenarios.

To alleviate the reliance on costly 3D annotations, recent research has explored self-supervised learning paradigms. For instance, the pioneering method S$^2$Hand~\cite{chen2021model} attempts to supervise hand reconstruction using noisy pseudo-labels produced by an off-the-shelf 2D joint detector, leveraging the discrepancy between input images and rendered outputs to drive network optimization. Building upon this paradigm, HaMuCo~\cite{zheng2023hamuco} introduces cross-view feature interaction and collaborative learning within the 2D visual space to progressively enhance estimation accuracy. 

However, pseudo-labels from 2D pose estimators inherently contain significant noise, and their indiscriminate use for supervision results in unstable network training. Moreover, existing methods fail to adequately exploit fine-grained spatial correlations and temporal dynamics, resulting in an insufficient capture of complex 3D spatial relationships and hand motions. To address the noise problem, we aim to preserve the uncertainty inherent in single-view estimation. By constructing a continuous, shared 3D space capable of capturing complex spatiotemporal relationships, we can leverage multi-view and temporal cues for effective pose disambiguation. In contrast to existing deterministic approaches, this strategy more effectively preserves 3D information, avoids interference caused by noisy 2D pseudo-labels, and facilitates the robust interaction of diverse pose hypotheses within a unified probabilistic feature space. 

To this end, we propose UST-Hand, an Uncertainty-aware Spatiotemporal point cloud interaction network for 3D hand pose estimation from synchronized and calibrated multi-view video sequences. Unlike previous methods that directly use deterministic pseudo-labels, UST-Hand effectively estimates uncertainty distributions and constructs a probabilistic point cloud space to comprehensively model hand motion. Specifically, UST-Hand employs a conditional normalizing flow to model the distribution of hand joints and sample diverse hypotheses for each view. This probabilistic formulation explicitly preserves uncertainty information, enabling the network to learn robust representations and significantly improving training stability by preventing noisy visual evidence from dominating the learning process. Subsequently, UST-Hand lifts these multi-view 2D hypotheses into a unified probabilistic 3D point cloud space for multi-view and temporal feature fusion. Furthermore, a Spatiotemporal Point Transformer (STPT) is designed as a dynamic interaction module to comprehensively capture fine-grained spatial correlations and temporal dynamics inherent in complex 3D hand structures and continuous motion trajectories. It leverages neighborhood interaction and iterative refinement to progressively improve hand pose estimation. By integrating the uncertainty-aware spatiotemporal disambiguation cue, UST-Hand effectively achieves high-quality 3D hand reconstruction, demonstrating exceptional robustness even under severely noisy pseudo-labels.

Experiments on three challenging multi-view hand pose estimation datasets—HanCo~\cite{zimmermann2021contrastive}, DexYCB~\cite{chao2021dexycb}, and OakInk~\cite{yang2022oakink}—show that UST-Hand outperforms existing self-supervised methods by up to 37.8\% in Mean Per Vertex Position Error (MPVPE). Remarkably, our approach delivers comprehensive improvements even under the challenging supervision of highly noisy pseudo-labels. More impressively, when leveraging pseudo-labels generated by the strongest supervised state-of-the-art (SOTA) models, our method still achieves significantly enhanced estimation accuracy. In summary, our contributions are listed as follows:
\begin{itemize}
    \item We propose UST-Hand, a self-supervised learning framework that employs conditional normalizing flows to model hand pose uncertainty distributions and sample diverse hypotheses. This enables robust learning under noisy pseudo-label supervision with significantly enhanced training stability.
    \item We introduce a unified probabilistic 3D point cloud feature space for multi-view and temporal interaction, and design a Spatiotemporal Point Transformer (STPT) that leverages neighborhood interaction and iterative refinement to comprehensively explore fine-grained spatial correlations and hand motion patterns.
    \item Extensive experiments on three challenging benchmarks demonstrate that UST-Hand achieves SOTA performance compared with existing self-supervised methods.
\end{itemize}

\section{Related Work}
\label{sec:related work}
\begin{figure*}
  \centering
  \includegraphics[width=1\linewidth]{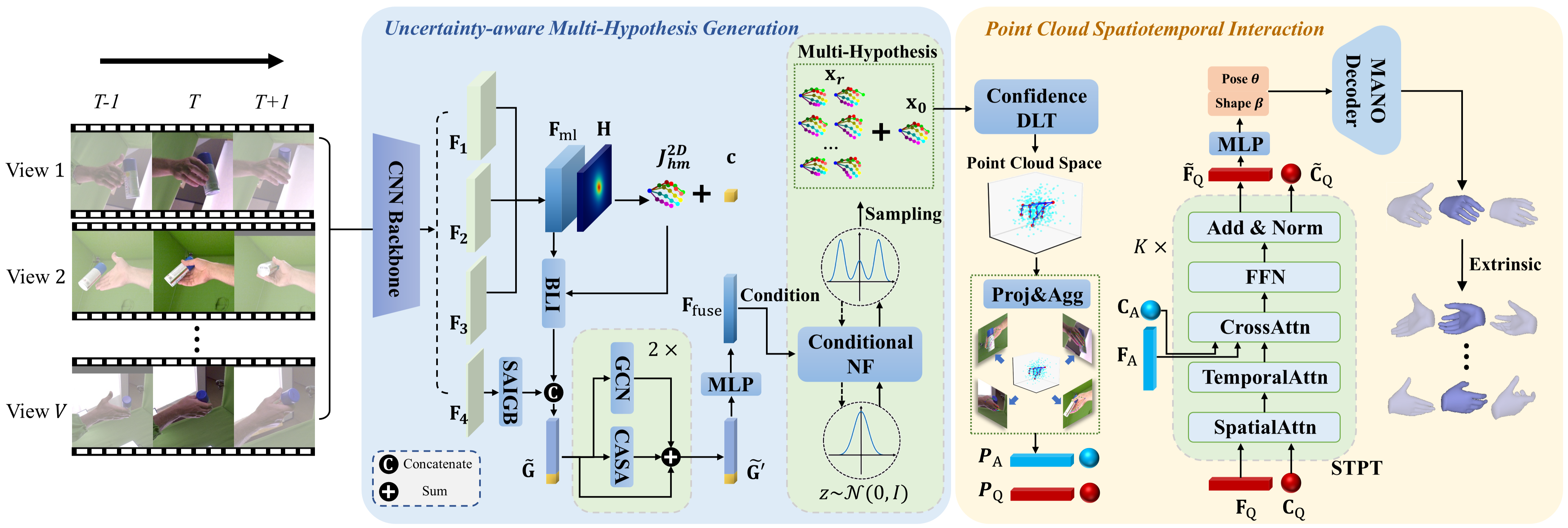}
  \caption{\textbf{Overview of the UST-Hand framework.} The reconstruction consists of two stages: (1) generating confidence-aware 2D features and sampling multi-view hypotheses via conditional normalizing flow (NF) to model uncertainty, and (2) lifting them into a unified probabilistic 3D point cloud space to explore spatiotemporal correlations via a Spatiotemporal Point Transformer (STPT).}
  \label{fig:flowchat}
\end{figure*}

\noindent\textbf{3D Hand Pose Estimation with Limited Supervision.}
While fully-supervised 3D hand pose estimation has advanced significantly with large-scale datasets~\cite{chao2021dexycb,garcia2018first,hampali2020honnotate,zimmermann2019freihand,zimmermann2021contrastive,kwon2021h2o,hampali2021ho,yang2022oakink,oh2023recovering,wang2025vihand}, acquiring precise 3D annotations remains prohibitively expensive. Consequently, recent efforts have shifted towards limited supervision paradigms~\cite{cai2018weakly,wan2019self,zhang2019end,spurr2020weakly,yang2021semihand,chen2021model,spurr2021self,chen2022mobrecon,chen2021temporal,ren2022dual,ren2022mining,zheng2023hamuco,ren2025rule}, leveraging synthetic data~\cite{chen2022mobrecon,yang2021semihand,wan2019self}, depth regularizers~\cite{cai2018weakly}, or biomechanical constraints~\cite{spurr2020weakly}. 

Notably, self-supervised approaches like S$^2$Hand~\cite{chen2021model} and HaMuCo~\cite{zheng2023hamuco} utilize 2D pseudo-labels from off-the-shelf detectors, with the latter incorporating multi-view consistency. However, these deterministic methods are highly susceptible to inherent pseudo-label noise and underutilize fine-grained spatiotemporal correlations. UST-Hand addresses these critical bottlenecks by modeling pose uncertainty distributions and establishing a probabilistic 3D point cloud space for robust spatiotemporal interaction.

\noindent\textbf{Multi-Hypothesis Pose Estimation.}
Given the inherent ambiguity in the 2D-to-3D lifting process of pose estimation, some studies have adopted the strategy of generating multiple hypotheses~\cite{ye2018occlusion,sharma2019monocular,li2019generating,li2021human,sengupta2021hierarchical,kolotouros2021probabilistic,oikarinen2021graphmdn,li2022mhformer,kundu2022uncertainty,chen2023mhentropy,bramlage2023plausible,biggs20203d,davoodnia2024upose3d,dwivedi2024poco,li2024hhmr,ge2025coarse,ma2025vmarker}. Early methods relied on random sampling~\cite{sengupta2021hierarchical}, multiple regression heads~\cite{biggs20203d}, or VAEs~\cite{ye2018occlusion,li2019generating}.

Recently, normalizing flows have emerged as a powerful tool for learning complex and ambiguous pose distributions~\cite{li2021human,dwivedi2024poco,davoodnia2024upose3d,chen2023mhentropy}. While methods like Ge~\etal~\cite{ge2025coarse} refine estimated distributions in discrete stages, our approach uniquely employs conditional normalizing flows to sample diverse view-specific hypotheses. We then lift these hypotheses into a unified probabilistic point cloud space, enabling stable and effective spatiotemporal disambiguation even under noisy pseudo-label supervision.

\noindent\textbf{Point Cloud Processing.}
As a compact spatial representation, point clouds preserve 3D spatial structure while avoiding redundant storage. Qi~\etal pioneered deep learning for point cloud processing, introducing PointNet~\cite{qi2017pointnet}, which paved the way for subsequent local convolution operators. More recently, Point Transformer~\cite{zhao2021point} uses vector attention for local neighboring points, while Point Cloud Transformer~\cite{guo2021pct} proposes a neighbor embedding strategy.

In 3D hand pose estimation, point-based networks have been utilized for point-wise regression~\cite{ge2018hand,ge2018point}, canonical alignment~\cite{chen2019so,deng2020weakly}, and multi-view frustum intersection (e.g., POEM~\cite{yang2023poem}). In contrast to these deterministic representations, our work explicitly constructs a probabilistic 3D point cloud space. Furthermore, we design a Spatiotemporal Point Transformer (STPT) that iteratively integrates distributional uncertainty to comprehensively capture fine-grained spatial correlations and temporal dynamics.

\section{Method}

UST-Hand is a self-supervised framework that achieves accurate and robust 3D hand pose estimation by explicitly modeling uncertainty distributions and leveraging spatiotemporal correlations under noisy pseudo-label supervision. The overview of UST-Hand is illustrated in Fig.~\ref{fig:flowchat}. To effectively decouple the noise inherent in 2D detections from the 3D lifting process, our framework operates in two synergistic stages: probabilistic 2D multi-hypothesis generation and 3D point cloud spatiotemporal interaction. Assuming a set of calibrated multi-view image sequences $\mathcal{I} = \{I_{v,t}\}_{v=1,t=1}^{V,T}$ captured from $V$ views across $T$ frames, UST-Hand aims to reconstruct 3D hand joints $\mathcal{J}^{\text{3D}} \in \mathbb{R}^{k \times 3}$ and vertices $\mathcal{V}^{\text{3D}} \in \mathbb{R}^{m \times 3}$, where $k=21$ and $m=778$, respectively. The supervisory signal relies exclusively on 2D pseudo-labels $\mathcal{J}^{\text{2D}}_{\text{pse}} \in \mathbb{R}^{k \times 2}$ generated by an off-the-shelf hand pose detector, making our uncertainty-aware formulation essential to prevent deterministic collapse and ensure stable convergence during model training.

\subsection{Heatmap-based Hand Joint Estimation}
To extract comprehensive visual representations from multi-view inputs, we adopt a pre-trained convolutional neural network (CNN) backbone built on residual structures. Hierarchical feature learning is realized through stacked residual blocks, producing multi-level feature maps $\mathcal{F} = \{\mathbf{F}_i\}_{i=1}^4$. Each block progressively halves the spatial dimension while increasing the channel dimension to capture abstract semantic information. We employ cascaded upsampling modules to restore deep features to higher spatial resolutions in a bottom-up manner, fusing them with corresponding hierarchical features from the encoding phase at each step. After restoring the resolution to $1/8$ of the original input size, a $1\times1$ convolutional layer transforms the multi-scale feature map $\mathbf{F}_{\text{ms}}$ into 2D likelihood heatmaps. The output heatmap $\mathbf{H}$ encodes the spatial probability distribution of each joint $\mathbf{p}_i \in \mathcal{J}^{\text{2D}}_{\text{hm}}$. We compute the joint location and corresponding confidence by:
\begin{equation}
    \mathbf{p}_i = \sum_{h_u} \sum_{h_v} (h_u, h_v) \cdot \mathbf{\tilde{H}}_i(h_u, h_v),
    \label{eq:Jhm}
\end{equation}
\begin{equation}
    \text{conf}_i = \max(\mathbf{H}_i),
    \label{eq:conf}
\end{equation}
where $h_u, h_v$ represent the horizontal and vertical coordinates of $\mathbf{H}_i$, and $\mathbf{\tilde{H}}$ is the normalized heatmap ensuring all values sum to $1$. The joint confidence $\mathbf{c}=\{ \text{conf}_i \}_{i=1}^{k}$ serves as a crucial indicator of pseudo-label quality, guiding subsequent joint-wise feature fusion, multi-view triangulation, and training loss weighting.

\subsection{Confidence-aware Feature Interaction}
To mitigate the adverse effects of noisy pseudo-labels, we design a confidence-aware mechanism that adaptively integrates reliability information into cross-view feature interactions. We first construct multi-view joint features through two complementary components: (1) spatial-aware joint features $\mathbf{G}_{\text{pose}}\!\in\!\mathbb{R}^{V\times k\times D_p}$ extracted from the high-resolution feature map $\mathbf{F}_4$ using SAIGB~\cite{zheng2021sar}, providing structural guidance across all views; (2) joint-aligned local features $\mathbf{G}_{\text{jaf}}\!\in\!\mathbb{R}^{V\times k\times D_a}$ obtained via bilinear interpolation (BLI) on $\mathcal{J}^{\text{2D}}_{\text{hm}}$ and $\mathbf{F}_{\text{ms}}$. Consequently, the initial graph is formed through feature concatenation:
\begin{equation}
    \mathbf{G}_{\text{init}}=\big[\mathbf{G}_{\text{pose}}\ \Vert\ \mathbf{G}_{\text{jaf}}\big] \!\in\!\mathbb{R}^{V\times k\times D},\quad D=D_p+D_a.
    \label{eq:finit}
\end{equation}

To explicitly incorporate uncertainty awareness, we propose a confidence-aware self-attention (CASA) mechanism. Specifically, we augment the initial graph by concatenating it with joint confidence and obtain $\tilde{\mathbf{G}}=\big[\mathbf{G}_{\text{init}}\ \Vert\ \mathbf{c}\big]$, then we compute the self-attention as:
\begin{equation}
    \mathbf{Q}=W_q\,\tilde{\mathbf{G}},\quad 
    \mathbf{K}=W_k\,\tilde{\mathbf{G}},\quad 
    \mathbf{V}=W_v\,\tilde{\mathbf{G}},
    \label{eq:qkv}
\end{equation}
\begin{equation}
    \text{Attn}=\text{softmax}\!\left(\frac{\mathbf{Q}\mathbf{K}^{\top}}{\sqrt{d}}\right)\mathbf{V}.
    \label{eq:attn}
\end{equation}

By embedding confidence into $\mathbf{Q}/\mathbf{K}/\mathbf{V}$, tokens associated with low-quality pseudo-labels naturally yield smaller query/key magnitudes and reduced mutual affinity, thereby suppressing their influence. After passing through alternating adaptive-GCN and CASA layers, $\tilde{\mathbf{G}}$ evolves into a robust cross-view interactive graph $\tilde{\mathbf{G}'}$. Finally, the fused feature $\mathbf{F}_\text{fuse}$ is obtained through an MLP, serving as the comprehensive 2D visual representation.

\subsection{Uncertainty-Aware Multi-Hypothesis Generation}
Unlike previous methods that rely on deterministic pseudo-labels, UST-Hand explicitly models the uncertainty distribution of hand poses to generate diverse hypotheses. We employ a conditional normalizing flow model, RealNVP~\cite{dinh2016density}, which learns a bijective mapping between a latent prior $\mathbf{z} \sim \mathcal{N}(\mathbf{0}, \mathbf{I})$ and 2D joint positions $\mathbf{x}$, conditioned on the cross-view features $\mathbf{F}_{\text{fuse}}$:
\begin{equation}
    \mathbf{x} = f_{\theta}(\mathbf{z}; \mathbf{F}_{\text{fuse}}), \quad \mathbf{z} = f_{\theta}^{-1}(\mathbf{x}; \mathbf{F}_{\text{fuse}}),
    \label{eq:flow}
\end{equation}
where $f_{\theta}$ denotes the invertible transformation parameterized by $\theta$. The conditional distribution is:
\begin{equation}
    \hat{p}(\mathbf{x}|\mathbf{F}_{\text{fuse}}) = p(\mathbf{z}) \left| \det \frac{\partial f_{\theta}(\mathbf{z}, \mathbf{F}_{\text{fuse}})}{\partial \mathbf{z}} \right|^{-1}.
    \label{eq:log}
\end{equation}

During training, we utilize the inverse transformation $f_{\theta}^{-1}$ to map pseudo-labels into the latent space and maximize the log-likelihood $\log \hat{p}(\mathbf{x}_{\text{pse}}|\mathbf{F}_{\text{fuse}})$. During inference, we sample $M$ instances $\mathbf{z}_r$ from the prior distribution and one key instance $\mathbf{z}_0$ from the mean, generating $M + 1$ hypotheses $\{\mathbf{x}_0; \mathbf{x}_r\}$ through $f_{\theta}$. Conceptually, rather than applying standard geometric constraints (e.g., RANSAC), the normalizing flow predicts a soft, continuous reliability distribution for each view. This probabilistic formulation ensures that partially occluded views are down-weighted but still contribute, while unreliable signals are effectively suppressed. This explicitly prevents noisy evidence from dominating the subsequent attention-based aggregation, leading to significantly enhanced training stability.

\subsection{Unified Probabilistic 3D Point Cloud Space}
To facilitate comprehensive spatiotemporal reasoning, we construct a unified probabilistic 3D point cloud space that aggregates multi-view information. We perform confidence-weighted Direct Linear Transform (DLT) triangulation on the sampled multi-hypothesis keypoints $\{\mathbf{x}_0; \mathbf{x}_r\}$, generating a 3D point cloud $\mathcal{P} = \{P_{\text{A}}; P_{\text{Q}}\}$ in world coordinates. Here, the anchor point cloud $P_{\text{A}}$, derived from random samples $\mathbf{x}_r$, is fixed to preserve distributional uncertainty, while the query point cloud $P_{\text{Q}}$, derived from the mode $\mathbf{x}_0$, serves as the refinement target that dynamically aggregates information from $P_{\text{A}}$.

To embed visual context into this geometric space, we augment the multi-scale feature maps $\mathbf{F}_{\text{ms}}$ with 2D spatial sine positional encoding to enhance multi-view awareness:
\begin{equation}
    \mathbf{F}_{\text{ms}} = \mathbf{F}_{\text{ms}} + \mathbf{F}_{\text{PE}}.
    \label{eq:emb}
\end{equation}

We project each 3D point to image coordinates $(u_p, v_p)$ across all views using camera parameters, and gather visual features $\{\mathbf{F}_i\}_{i=1}^V$ via BLI. These multi-view features are fused through a bottleneck network to produce point-wise features $\mathbf{F}_{\text{P}}$. This establishes a unified probabilistic 3D point cloud space jointly representing geometry and visual cues, enabling effective spatiotemporal interaction.

\subsection{Spatiotemporal Point Transformer for Iterative Refinement}
We formulate the anchor point cloud as $P_{\text{A}} = (\mathbf{C}_\text{A}, \mathbf{F}_\text{A})$ and the query point cloud as $P_{\text{Q}} = (\mathbf{C}_\text{Q}, \mathbf{F}_\text{Q})$, where $\mathbf{C}$ and $\mathbf{F}$ denote coordinates and features, respectively. To comprehensively explore fine-grained spatial correlations and temporal motion patterns, we design STPT that iteratively refines $P_{\text{Q}}$ through a dual-stage attention mechanism.

\paragraph{Spatiotemporal Self-Attention.}
Our dual-stage attention mechanism processes query sequences of shape $(B \times T, J, D)$, where $B$, $T$, and $J$ represent batch size, temporal window size, and the number of joints, respectively. We first model intra-frame geometric relationships via neighborhood-aware spatial attention. For each query point $i$, we retrieve its $k$-nearest neighbors and compute:
\begin{equation}
    \mathbf{F}_\text{Q}^s = \text{SpatialAttn}(\mathbf{C}_{\text{Q},i} - \mathbf{C}_{\text{Q},j}),
    \label{eq:sa}
\end{equation}
where relative position encoding $\delta_s = \theta(\mathbf{C}_{\text{Q},i} - \mathbf{C}_{\text{Q},j})$ encodes the geometric context between point $i$ and its neighborhood $j$, capturing local structural dependencies.

Following spatial attention, features are reshaped to $(B, T, J, D)$ to enable temporal modeling. We inject learnable frame-wise positional embeddings and apply temporal self-attention along the time axis for each hand joint to captures hand motion patterns across temporal sequences:
\begin{equation}
    \mathbf{F}_\text{Q}^{st} = \text{TemporalAttn}(\mathbf{F}_\text{Q}^s).
    \label{eq:ta}
\end{equation}

\paragraph{Cross-Attention and Iterative Position Refinement.}
We establish cross-set correspondence between temporally-aware query features $\mathbf{F}_\text{Q}^{st}$ and the anchor point cloud $P_{\text{A}}$ to leverage distributional uncertainty:
\begin{equation}
    \tilde{\mathbf{F}}_{\text{Q},i} = \text{CrossAttn}(\mathbf{F}_{\text{Q},i}^{st}, \{\mathbf{F}_{\text{A},j}\}_{j \in \mathcal{X}_i}),
    \label{eq:cr}
\end{equation}
where CrossAttn denotes the cross-attention operation with position encoding $\delta = \theta(\mathbf{C}_{\text{Q},i} - \mathbf{C}_{\text{A},j})$ over the $k$-nearest neighbors $\mathcal{X}_i$. The refinement can be formulated as:
\begin{equation}
    \tilde{\mathbf{C}}_{\text{Q},i} = \mathbf{C}_{\text{Q},i} + \text{FFN}(\tilde{\mathbf{F}}_{\text{Q},i}).
    \label{eq:Qc}
\end{equation}

Through $K$ refinement iterations, STPT progressively improves pose estimation by integrating spatial neighborhood information, temporal motion cues, and distributional uncertainty from the anchor point cloud.

\paragraph{Parametric Output.}
Upon completing $K$ refinement iterations, we obtain final query representations $\tilde{\mathbf{F}}_{\text{Q}}$ and refined coordinates $\tilde{\mathbf{C}}_{\text{Q}}$. We employ the MANO model~\cite{romero2022embodied} to generate explicit 3D hand mesh $\mathcal{V}^{\text{3D}}$ from the refined joint positions, producing the final pose estimation.

\subsection{Loss Functions}
During the training phase, we strictly use 2D joint locations predicted by an offline detector as pseudo-labels. 

A custom mean squared error (MSE) loss is adopted to supervise the predicted heatmap $\mathcal{H}_{\text{pred}}$ by:
\begin{equation}
    \mathcal{L}_{\text{hmap}} = \frac{1}{k}\sum_{i=1}^k(\mathcal{H}_{\text{pred}, i} - \mathcal{H}_{\text{pse}, i})^2,
    \label{eq:hmap}
\end{equation}
where the pseudo-label heatmap $\mathcal{H}_{\text{pse}, i}$ is generated from $\mathcal{J}^{\text{2D}}_{\text{pse}, i}$ through a 2D Gaussian function with $\sigma=2$.

We further supervise the 2D joint locations decoded from the predicted heatmap by:
\begin{equation}
    \mathcal{L}_{\text{hm2d}} = \frac{1}{k}\sum_{i=1}^k\left(\mathcal{J}^{\text{2D}}_{\text{hm}, i} - \mathcal{J}^{\text{2D}}_{\text{pse}, i} \right)^2.
    \label{eq:hm2d}
\end{equation}

We utilize a negative log-likelihood loss to control the uncertainty distribution of multi-hypothesis sampling:
\begin{equation}
    \mathcal{L}_{nll} = -\log \hat{p}(\mathbf{x}|\mathbf{F}_{\text{fuse}}).
    \label{eq:nll}
\end{equation}

For the refined query points $\mathcal{J}_{\text{query}}^{\text{3D}}$ and final hand joint $\mathcal{J}_{\text{mano}}^{\text{3D}}$, we project them onto the pixel space of each view to get their 2D positions $\mathcal{J}_{\text{query}}^{\text{2D}}$ and $\mathcal{J}_{\text{mano}}^{\text{2D}}$. We then supervise them with an $L_2$ distance loss using 2D pseudo-labels:
\begin{equation}
\begin{split}
    \mathcal{L}_{\text{proj2d}} = \frac{1}{N}\sum_{j=1}^N & \left( \frac{1}{k}\sum_{i=1}^k \text{conf}_i\left\|\mathcal{J}^{\text{2D}}_{\text{query}, i} - \mathcal{J}^{\text{2D}}_{\text{pse}, i} \right\|_2^2 \right) \\
    + \frac{1}{k}& \sum_{i=1}^k \text{conf}_i\left\|\mathcal{J}^{\text{2D}}_{\text{mano}, i} - \mathcal{J}^{\text{2D}}_{\text{pse}, i} \right\|_2^2.
    \label{eq:proj2d}
\end{split}
\end{equation}

The final loss can be formulated as:
\begin{equation}
    \mathcal{L} = \lambda_0\mathcal{L}_{\text{hmap}} + \lambda_1\mathcal{L}_{\text{hm2d}} + \lambda_2\mathcal{L}_{nll} + \lambda_3\mathcal{L}_{\text{proj2d}},
    \label{eq:total}
\end{equation}
where $\lambda_0, \lambda_1, \lambda_2, \lambda_3$ are the tradeoff parameters to balance the loss scale. Specifically, we set $\lambda_0=0.001$, $\lambda_1=10$, $\lambda_2=0.1$, and $\lambda_3=10$.


\section{Experiment}
\label{sec:exp}

\subsection{Datasets and Evaluation Metrics}
\label{sec:datasets}
\noindent\textbf{HanCo.} HanCo~\cite{zimmermann2021contrastive} provides multi-view hand image sequences captured at 5Hz by 8 synchronized cameras, offering crucial temporal and spatial consistency for unsupervised learning. It contains 107,538 temporal instances, yielding 860,304 RGB images in total.

\noindent\textbf{DexYCB-MV.} DexYCB~\cite{chao2021dexycb} captures hand-object manipulations from 8 viewpoints. We adopt the official 'S0' partition and retain only right-hand observations. By aggregating synchronized multi-view images per temporal instance, we construct DexYCB-MV, comprising 25,387 training, 1,412 validation, and 4,951 testing multi-view samples.

\noindent\textbf{OakInk-MV.}
OakInk~\cite{yang2022oakink} provides 230K images from a 4-camera setup. Following the official 'SP2' protocol, we construct OakInk-MV by synchronizing images across all views. For sequences involving two hands ($\sim$25\%), each hand is processed independently. The dataset yields 58,692 m.v. training samples and 19,909 testing samples.

\noindent\textbf{Metrics.} We evaluate our model using Mean Per Joint Position Error (MPJPE) and Mean Per Vertex Position Error (MPVPE), along with their Procrustes-aligned variants (PA-J/PA-V)~\cite{gower1975generalized}. We also report the F-Score~\cite{chen2021model} at 5mm and 15mm, and the Area Under the Curve (AUC) for the Percentage of Correct Keypoints (PCK) across 100 uniformly spaced thresholds ranging from 0 to 50mm.

\begin{table*}[t]
\caption{Quantitative results on three multi-view datasets, HanCo, DexYCB-MV, OakInk-MV. The AUC-V and AUC-J are computed on MPVPE and MPJPE respectively, with the thresholds setting to 0-50~mm for all three datasets.}
\centering
\resizebox{\linewidth}{!}{
\begin{tabular}{c|c|l|c|c|c|c|c|c|c|c}
\toprule
\multicolumn{2}{c|}{\multirow{2}{*}{}} & \multirow{2}{*}{\textbf{Methods}} & \multicolumn{5}{c|}{\textbf{Hand vertices}} & \multicolumn{3}{c}{\textbf{Hand joints}} \\
\cline{4-11}
\multicolumn{2}{c|}{} &  & \textbf{MPVPE}$\downarrow$ & \textbf{PA-V}$\downarrow$ & \textbf{AUC-V}$\uparrow$ & \textbf{F@5}$\uparrow$ & \textbf{F@15}$\uparrow$ & \textbf{MPJPE}$\downarrow$ & \textbf{PA-J}$\downarrow$ & \textbf{AUC-J}$\uparrow$ \\
\midrule
\multirow{6}{*}{\rotatebox{90}{\shortstack{HanCo \\ (8 views)}}} & 1 & UST-Hand & \textbf{5.82} & \textbf{4.13} & \textbf{0.884} & \textbf{0.749} & \textbf{0.999} & \textbf{5.19} & \textbf{3.50} & \textbf{0.897} \\
& 2 & HaMuCo~\cite{zheng2023hamuco} & 9.35 & 5.38 & 0,813 & 0.581 & 0.974 & 8.73 & 4.77 & 0.826 \\
& 3 & Wilor~\cite{potamias2025wilor} & 12.70 & 5.49 & 0.746 & 0.432 & 0.937 & 11.80 & 5.95 & 0.765 \\
& 4 & UST-Hand w/o hmap. & 6.42 & 4.33 & 0.872 & 0.716 & 0.998 & 5.80 & 3.69 & 0.884 \\
& 5 & UST-Hand w/o proj. & 6.14 & 4.25 & 0.877 & 0.716 & 0.998 & 5.64 & 3.80 & 0.887 \\
& 6 & UST-Hand w/o STPT. & 6.05 & 4.23 & 0.879 & 0.712 & 0.998 & 5.53 & 3.66 & 0.890 \\
\midrule
\multirow{6}{*}{\rotatebox{90}{\shortstack{DexYCB-MV \\ (8 views)}}} & 7 & UST-Hand & \textbf{8.16} & \textbf{4.81} & \textbf{0.837} & \textbf{0.609} & \textbf{0.982} & \textbf{7.84} & \textbf{5.31} & \textbf{0.843} \\
& 8 & HaMuCo~\cite{zheng2023hamuco} & 9.54 & 5.32 & 0.810 & 0.539 & 0.963 & 9.25 & 5.71 & 0.816 \\
& 9 & Wilor~\cite{potamias2025wilor} & 10.83 & 5.88 & 0.787 & 0.524 & 0.936 & 10.52 & 6.08 & 0.793 \\
& 10 & UST-Hand w/o hmap. & 8.94 & 5.43 & 0.825 & 0.554 & 0.973 & 8.76 & 5.90 & 0.825 \\
& 11 & UST-Hand w/o proj. & 8.70 & 5.15 & 0.826 & 0.574 & 0.979 & 8.40 & 5.59 & 0.832 \\
& 12 & UST-Hand w/o STPT. & 8.78 & 5.17 & 0.825 & 0.571 & 0.976 & 8.46 & 5.69 & 0.831 \\
\midrule
\multirow{6}{*}{\rotatebox{90}{\shortstack{OakInk-MV \\ (4 views)}}} & 13 & UST-Hand & \textbf{10.02} & \textbf{5.72} & \textbf{0.800} & \textbf{0.511} & \textbf{0.982} & \textbf{8.70} & \textbf{5.58} & \textbf{0.827} \\
& 14 & HaMuCo~\cite{zheng2023hamuco} & 13.04 & 7.12 & 0.751 & 0.433 & 0.937 & 12.03 & 6.90 & 0.770 \\
& 15 & Wilor~\cite{potamias2025wilor} & 15.66 & 7.37 & 0.707 & 0.429 & 0.887 & 15.08 & 7.49 & 0.718 \\
& 16 & UST-Hand w/o hmap. & 11.75 & 6.18 & 0.778 & 0.487 & 0.977 & 10.08 & 5.99 & 0.803 \\
& 17 & UST-Hand w/o proj. & 10.41 & 5.77 & 0.793 & 0.501 & 0.981 & 9.08 & 5.69 & 0.820 \\
& 18 & UST-Hand w/o STPT. & 10.77 & 5.86 & 0.789 & 0.490 & 0.980 & 9.01 & 5.74 & 0.818 \\
\bottomrule
\end{tabular}
}
\label{tab:quantitative}
\end{table*}

\subsection{Implementation Details}
The implementation is based on PyTorch and executed on dual 24G NVIDIA RTX 4090 GPUs. Training is conducted over 30 epochs using a batch size of 8. Model parameters are optimized via the Adam algorithm, starting with a learning rate of $3 \times 10^{-4}$ which undergoes a tenfold reduction at epoch 20. Across all multi-view experimental settings, ResNet34 \cite{he2016deep} serves as the backbone network, with parameters initialized from ImageNet pre-training weights. For temporal processing, we construct a sliding window of length 5 with 1 stride, positioned centrally around each frame in the video sequence. For frames lacking sufficient context (at sequence boundaries), we apply replication padding by duplicating the closest available frames. The temporal and batch axes are merged prior to applying temporal self-attention mechanisms, then separated during attention computation to effectively model hand movement dynamics. During training, we employ conventional data augmentation, including random center offset (within $\pm5\%$), scaling ($\pm6\%$), color jittering ($\pm30\%$), and rotation ($\pm10^\circ$). To preserve consistency across time, all frames within the same numbers receive identical augmentation.

\subsection{Quantitative Evaluation}
As shown in Tab.~\ref{tab:quantitative}, our UST-Hand method (rows \#1, \#7, \#13) achieves state-of-the-art performance across all three datasets under the same multi-view settings. Compared to HaMuCo~\cite{zheng2023hamuco} (rows \#2, \#8, \#14), the previous state-of-the-art self-supervised approach, UST-Hand significantly reduces MPVPE by 37.8\% on HanCo (9.35mm$\rightarrow$5.82mm), 14.5\% on DexYCB-MV (9.54mm$\rightarrow$8.16mm), and 23.2\% on OakInk-MV (13.04mm$\rightarrow$10.02mm). Similar substantial improvements are observed in MPJPE and other evaluation metrics, with AUC-V gains of 16.2\%, 3.3\%, and 6.5\% respectively. To the best of our knowledge, HaMuCo remains the only other method that exploits multi-view data for self-supervised hand pose estimation.

We further compare against Wilor~\cite{potamias2025wilor} (rows \#3, \#9, \#15), a SOTA offline monocular hand pose detector for producing pseudo labels. Wilor is selected for two key reasons: first, its superior occlusion-handling capability makes it a rigorous benchmark for performance evaluation; second, as an advanced method capable of generating 3D hand meshes, it provides direct and meaningful reference for assessing 3D reconstruction quality. Experimental results show that our method substantially outperforms Wilor in 3D pose estimation, demonstrating the effectiveness of our network architecture and training strategy.

\subsection{Ablation Study}
To validate the contribution of each component, we conduct comprehensive ablation studies on all three datasets, with results presented in rows \#4-6, \#10-12, and \#16-18 of Tab.~\ref{tab:quantitative}. See the supplement for more details.

\noindent\textbf{Heatmap Module.} The heatmap generation in our first stage provides essential coordinate priors for 2D visual feature interaction and produces joint confidence scores that reflect pseudo-label reliability. These confidence scores guide feature interaction, algebraic triangulation, and loss weighting throughout the network. Removing heatmaps (rows \#4, \#10, \#16) and setting all confidences to 1 leads to MPVPE increases of 10.3\%, 9.6\%, and 17.3\% across the three datasets, confirming that confidence-weighted initialization is crucial for handling noisy pseudo-labels.
\begin{table}
\caption{Quantitative results (mm) on the HanCo dataset.}
\centering
\small
\setlength{\tabcolsep}{10pt}  
\renewcommand{\arraystretch}{1.1}
\begin{tabular}{clcc}
\toprule
\textbf{View} & \textbf{Method} & \textbf{MPJPE} $\downarrow$ & \textbf{PA-J} $\downarrow$ \\
\midrule
\multirow{2}{*}{2} & HaMuCo & 10.14 & 5.92 \\
 & UST-Hand\_t1 & \textbf{7.18} & \textbf{4.84} \\
 \midrule
\multirow{2}{*}{4} & HaMuCo & 9.60 & 5.53 \\
 & UST-Hand\_t1 & \textbf{6.01} & \textbf{3.93} \\
 \midrule
\multirow{2}{*}{6} & HaMuCo & 8.92 & 5.11 \\
 & UST-Hand\_t1 & \textbf{5.67} & \textbf{3.67} \\
 \midrule
\multirow{2}{*}{8} & HaMuCo & 8.73 & 4.77 \\
 & UST-Hand\_t1 & \textbf{5.38} & \textbf{3.55} \\
\bottomrule
\end{tabular}
\label{tab:sparse_view}
\end{table}
\begin{table}
\caption{Multi-view inference results under the supervision of 2D pseudo-labels of varying quality on the HanCo dataset.}
\centering
\small
\setlength{\tabcolsep}{10pt}  
\renewcommand{\arraystretch}{1.1}
\begin{tabular}{lcc}  
\toprule
\textbf{Method} & \textbf{MPJPE} $\downarrow$ & \textbf{PA-J} $\downarrow$ \\
\midrule
\multicolumn{3}{l}{\emph{Self-Supervised (OpenPose Pseudo-labels):}} \\
EpipolarTrans \cite{he2020epipolar}    & 11.2 & 9.0 \\
LT-Algebraic \cite{iskakov2019learnable}     & 10.3 & 7.8 \\
LT-Volumetric \cite{iskakov2019learnable}    & 10.6 & 8.0 \\
HaMuCo \cite{zheng2023hamuco}             & 8.8 & 5.3 \\
\rowcolor{oursblue!50}  
UST-Hand         & \textbf{7.5} & \textbf{5.0} \\
\midrule
\multicolumn{3}{l}{\emph{Self-Supervised (Wilor Pseudo-labels ):}} \\
EpipolarTrans \cite{he2020epipolar}    & 9.3 & 7.9 \\
LT-Algebraic \cite{iskakov2019learnable}     & 8.5 & 5.7 \\
LT-Volumetric \cite{iskakov2019learnable}    & 8.4 & 5.8 \\
HaMuCo \cite{zheng2023hamuco}             & 8.7 & 4.6 \\
\rowcolor{oursblue!50}
UST-Hand         & \textbf{5.2} & \textbf{3.5} \\
\midrule
\multicolumn{3}{l}{\emph{Fully-Supervised (2D Ground-truth)}} \\
EpipolarTrans \cite{he2020epipolar}    & 6.2 & 4.2 \\
LT-Algebraic \cite{iskakov2019learnable}     & 5.5 & 3.6 \\
LT-Volumetric \cite{iskakov2019learnable}    & 5.8 & 3.6 \\
HaMuCo  \cite{zheng2023hamuco}      & 6.0 & 3.2 \\
\rowcolor{oursblue!50}
UST-Hand         & \textbf{3.7} & \textbf{2.4} \\
\bottomrule
\end{tabular}
\label{tab:pseudo_label}
\end{table}

\noindent\textbf{Projection Fusion Module.} This component connects the 3D probabilistic point cloud with 2D visual observations by fusing the retrieved 2D features into comprehensive 3D visual perception. Ablating this module (rows \#5, \#11, \#17) and simply concatenating 2D features degrades MPVPE by 5.5\%, 6.6\%, and 3.9\%, highlighting the importance of explicitly establishing geometry-vision correspondence for accurate pose refinement.

\noindent\textbf{STPT Module.} As the key component of our second stage, STPT performs information interaction over the point cloud to continuously refine hand pose features by exploring complex spatial-temporal relationships across multiple views and frames. Without STPT (rows \#6, \#12, \#18), directly estimating MANO parameters from query features causes MPVPE to increase by 4.0\%, 7.6\%, and 7.5\%. This demonstrates the importance of explicit reasoning hand pose over a unified space through spatiotemporal interaction.
\begin{figure}
  \centering
  \includegraphics[width=1\linewidth]{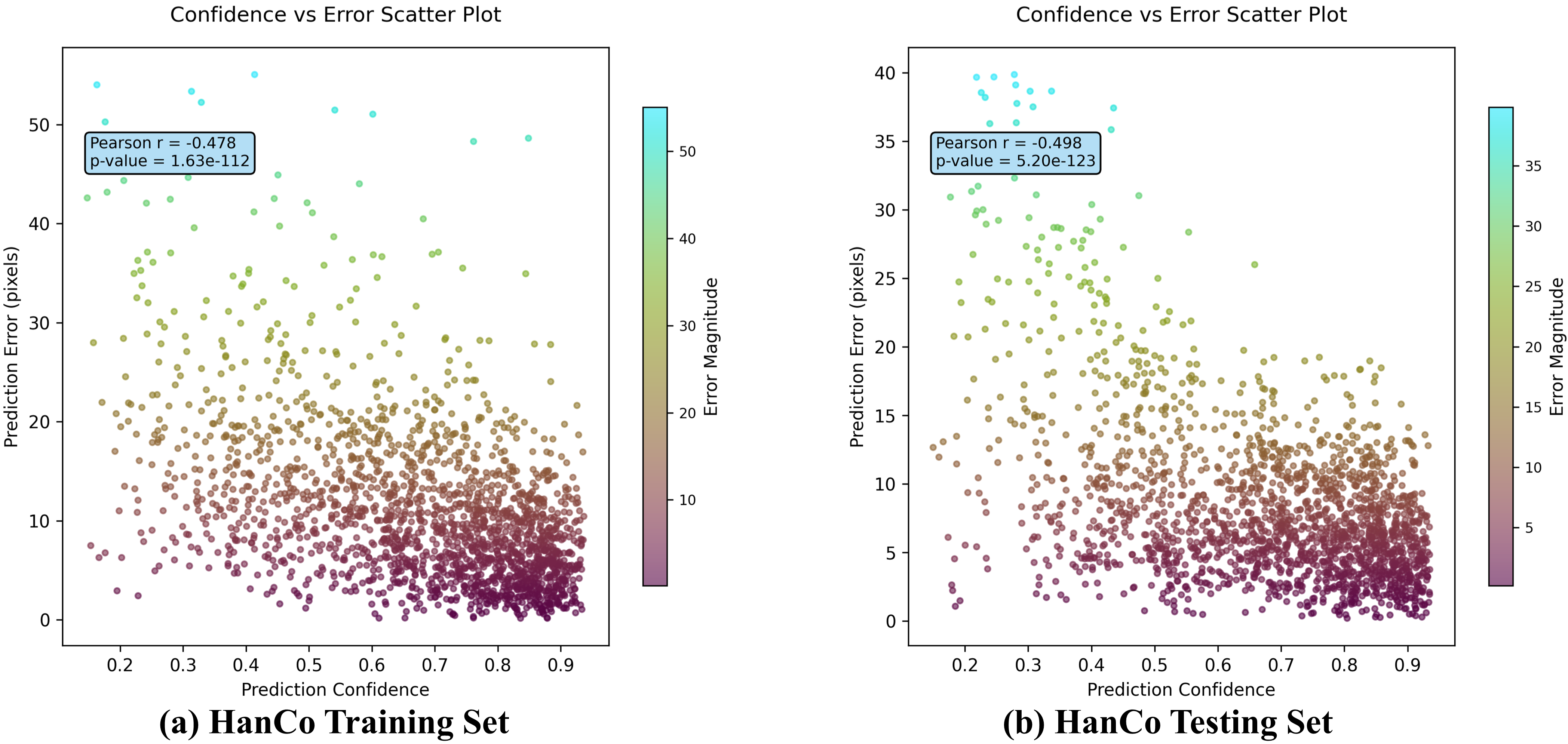}
  \caption{The relationship between confidence and joints error.}
  \label{fig:conf}
\end{figure}
\begin{figure}
  \centering
  \includegraphics[width=1\linewidth]{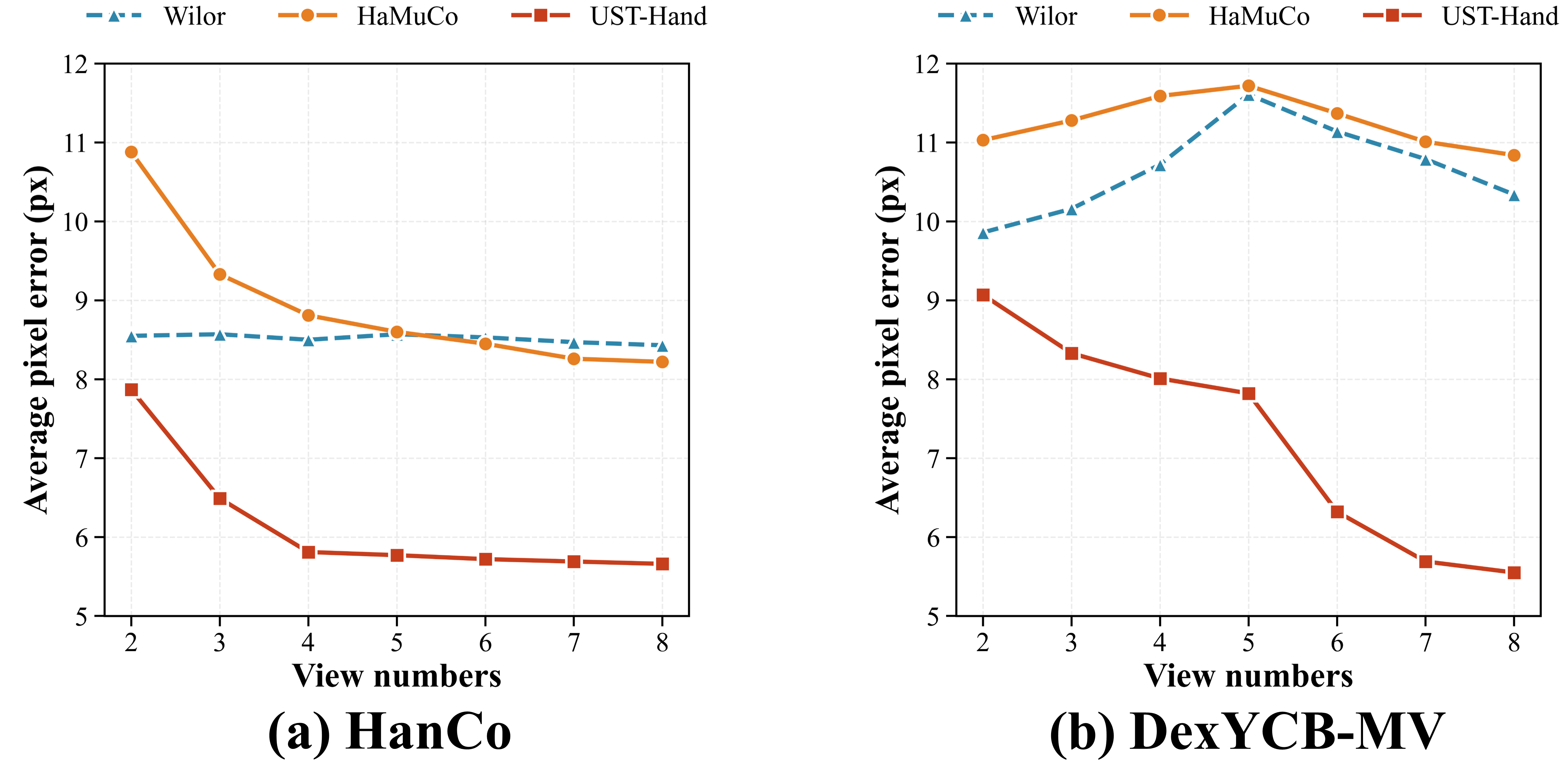}
  \caption{Average 2D pixel error under different view settings.}
  \label{fig:2derr}
\end{figure}
\begin{figure}
  \centering
  \includegraphics[width=1\linewidth]{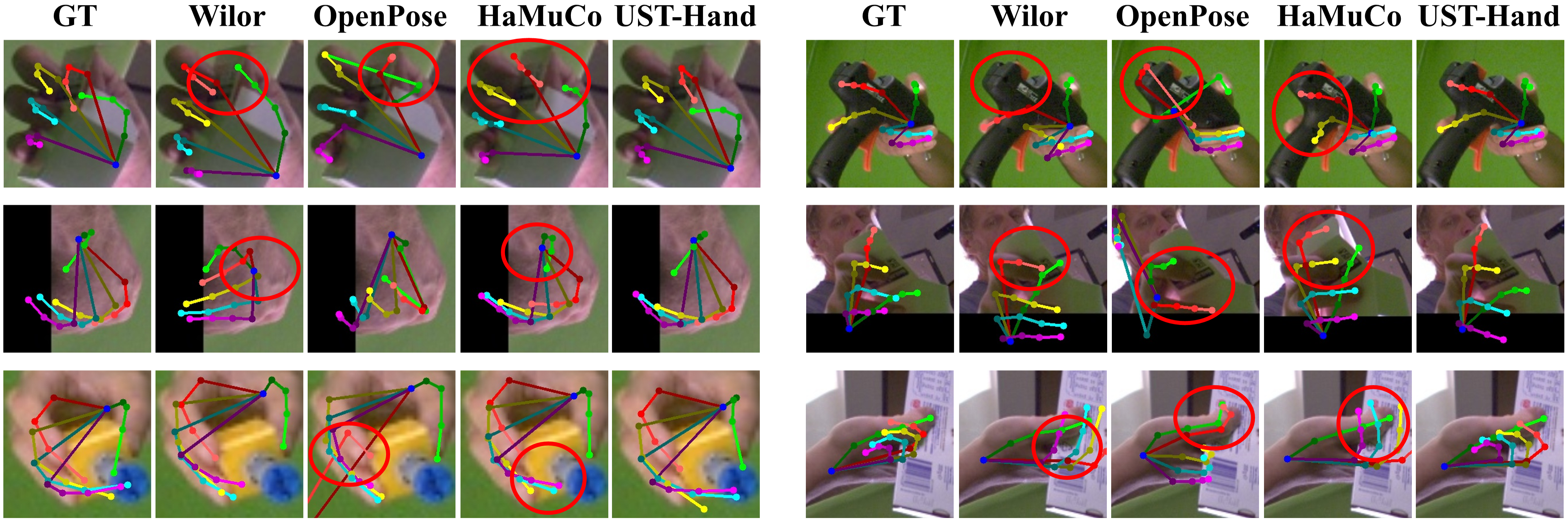}
  \caption{visualization of 2D joint predictions (overlaid on the images) among the ground-truth (GT), offline hand pose detectors (Wilor and OpenPose), the compared method HaMuCo, and UST-Hand on the HanCo dataset. Regions with significant prediction errors are highlighted with red circles.}
  \label{fig:vis2D}
\end{figure}

\noindent\textbf{Sparse-view Robustness.} To evaluate performance under limited camera setups, we compare HaMuCo against our single-frame variant (UST-Hand\_t1) across varying view densities on the HanCo dataset, using Wilor pseudo-labels. As shown in Tab.~\ref{tab:sparse_view}, UST-Hand\_t1 consistently achieves state-of-the-art accuracy under \emph{every} view configuration. Notably, even without relying on temporal cues, our method significantly outperforms HaMuCo in the standard 8-view setting and maintains striking superiority in highly sparse scenarios (yielding MPJPE improvements of 2.96, 3.59, 3.25, and 3.35\,mm under 2, 4, 6, and 8 views, respectively). While fewer views inevitably increase geometric ambiguity, our uncertainty-aware formulation effectively avoids deterministic collapse, demonstrating exceptional robustness and generalization across diverse multi-view setups.
\begin{figure*}
  \centering
  \includegraphics[width=1\linewidth]{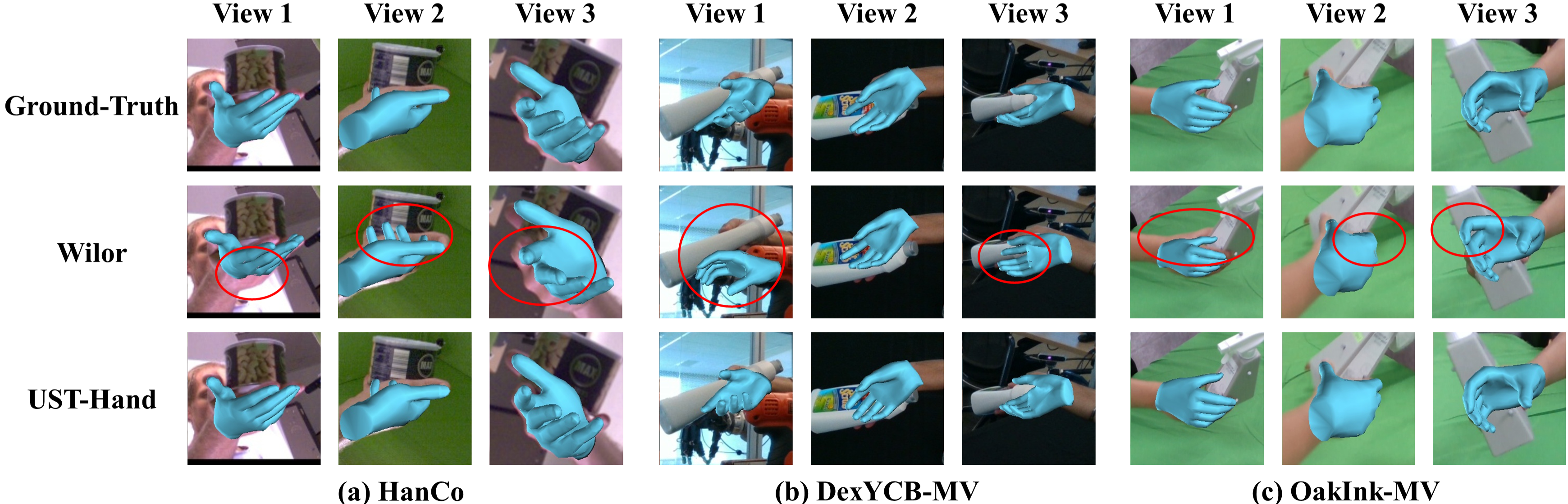}
  \caption{The 3D mesh visualization (overlaid in the images) between ground-truth, Wilor, and UST-Hand on (a) HanCo, (b) DexYCB-MV, and (c) OakInk-MV datasets. The regions with significant prediction errors have been circled in red.}
  \label{fig:vis3D}
\end{figure*}
\subsection{Robustness to Pseudo-Label Quality}
To evaluate robustness under varying pseudo-label quality, we conduct experiments with three supervision settings on HanCo, as shown in Tab.~\ref{tab:pseudo_label}. We compare against EpipolarTrans~\cite{he2020epipolar}, Learnable Triangulation (LT)~\cite{iskakov2019learnable}, and HaMuCo~\cite{zheng2023hamuco}, where EpipolarTrans and LT are adapted from fully-supervised multi-view human pose estimation to self-supervised settings. Unlike deterministic methods like HaMuCo that tend to overfit to pseudo-label noise, our uncertainty-aware formulation effectively denoises the 2D pseudo-labels. Using low-quality OpenPose detections that lack hand model constraints and exhibit large errors, UST-Hand achieves 7.5\,mm MPJPE, outperforming HaMuCo by 14.8\%. Notably, even when building upon Wilor with exceptional quality and occlusion robustness, UST-Hand still achieves substantial improvements with 5.3\,mm MPJPE, outperforming HaMuCo by 26.4\%. Under ground-truth 2D supervision exploring the upper bound, UST-Hand achieves 3.7\,mm MPJPE and 2.4\,mm PA-MPJPE, surpassing HaMuCo by 38.3\%. Across all settings, UST-Hand consistently achieves the best performance, proving our inherent robustness against noisy supervision and the capability to effectively exploit high-quality cues.
\subsection{Qualitative Result}
Fig.~\ref{fig:conf} presents the confidence visualization results of selected hand keypoints for UST-Hand during training and testing on the HanCo dataset. A significant inverse correlation between the predicted confidence scores and pose estimation errors is observed. This demonstrates that UST-Hand effectively learns meaningful keypoint confidences that accurately reflect prediction quality.

Fig.~\ref{fig:2derr} compares the average 2D pixel errors across varying view counts. Here, Wilor provides the noisy pseudo-labels for self-supervised training, exhibiting natural error fluctuations as views change. The curves demonstrate that UST-Hand consistently achieves lower errors than the Wilor baseline under all settings. In contrast, HaMuCo degrades severely under sparse views and merely approaches Wilor with dense views. This indicates a tendency for HaMuCo to overfit to noisy supervision, whereas UST-Hand overcomes this limitation, showcasing exceptional robustness.

Fig.~\ref{fig:vis2D} illustrates a qualitative comparison of 2D hand keypoint predictions on the HanCo dataset. Among the offline hand pose detectors, the 2D keypoints generated by OpenPose are highly susceptible to occlusion and image truncation, resulting in substantial noise. While Wilor significantly improves occlusion robustness and yields more plausible hand poses, substantial errors still remain. In contrast, by effectively leveraging geometric correction, our method demonstrates superior occlusion handling ability. UST-Hand not only outperforms both the offline detectors and HaMuCo, but also achieves 2D keypoint predictions that are most consistent with the ground truth.

Furthermore, Fig.~\ref{fig:vis3D} shows the visualization results of 3D hand meshes from 3 different views. The comparative results vividly demonstrate that our method improves the accuracy of 3D hand reconstruction. More qualitative results are provided in the Supplementary Material.



\section{Conclusion}
This work presents UST-Hand, a self-supervised framework for 3D hand pose estimation from synchronized multi-view video sequences, effectively eliminating costly 3D annotations. To overcome training instability caused by noisy pseudo-labels, UST-Hand employs conditional normalizing flows to explicitly model pose uncertainty. Diverse 2D hypotheses are lifted into a unified probabilistic 3D point cloud space, where a Spatiotemporal Point Transformer (STPT) iteratively refines poses. Extensive experiments on three challenging benchmarks demonstrate state-of-the-art performance. Our efficient model supports real-time multi-view and monocular inference, maintaining striking superiority across varying camera densities and pseudo-label qualities. Ultimately, this uncertainty-aware probabilistic paradigm offers a promising solution for general articulated object reconstruction under noisy supervision.

\noindent\textbf{Acknowledgements} This work was supported in part by the grants from the National Natural Science Foundation of China under Grant 62332019, the National Key Research and Development Program of China (2023YFF1203900, 2023YFF1203903), sponsored by the National Natural Science Foundation of China under Grants (62406039).



{
    \small
    \bibliographystyle{ieeenat_fullname}
    \bibliography{main}
}

\clearpage
\setcounter{page}{1}
\maketitlesupplementary

\section{Video Demo}
We provide sequential visualizations in the attached video to illustrate our method’s performance.

\section{Additional Ablation Study}
To further validate our approach, we conduct fine-grained ablation studies on the individual components within the Confidence-aware feature interaction module and the Spatiotemporal Point Transformer (STPT). As shown in~\cref{tab:addition_ablation}, in the Confidence-aware feature interaction module, removing the adaptive-GCN or CASA mechanism leads to performance degradation, demonstrating that both structural topology modeling and confidence-aware feature aggregation contribute to robust cross-view feature interaction. For the STPT module, temporal attention shows the largest impact, highlighting the importance of exploiting temporal consistency for video-based hand pose estimation. These results confirm that each component plays an indispensable role in achieving accurate pose refinement.
\begin{table}[htbp]
\caption{Additional Ablation Study of UST-Hand on DexYCB-MV dataset. We report the MPVPE (mm), PA-V (mm), MPJPE (mm), PA-J (mm).}
\label{tab:addition_ablation}
\begin{tabular}{c|cccc}
\hline
Method           & MPVPE & PA-V & MPJPE & PA-J \\ \hline
UST-Hand         & 8.16  & 4.81 & 7.84  & 5.31 \\ \hline
w/o GCN          & 8.22  & 4.83 & 7.90  & 5.41 \\
w/o CASA         & 8.28  & 4.87 & 7.95  & 5.43 \\ \hline
w/o SpatialAttn  & 8.46  & 5.02 & 8.18  & 5.58 \\
w/o TemporalAttn & 8.55  & 5.09 & 8.20  & 5.64 \\
w/o CrossAttn    & 8.50  & 4.99 & 8.15  & 5.59 \\ \hline
\end{tabular}
\end{table}
\section{Model Analysis}
\begin{table}[ht]
\centering
\caption{Performance of UST-Hand on varying temporal length, the number of STPT blocks and cameras in the multi-view setting. The best results are highlighted in \textbf{bold}.}
\label{tab:UST-Hand_performance}
\begin{tabular}{c|c|cccc}
\hline
& Exp & MPVPE & PA-V & MPJPE & PA-J \\
\hline
\multirow{8}{*}{\rotatebox{90}{HanCo}} 
& t1 & 5.94 & 4.15 & 5.38 & 3.55 \\
& t3 & 5.91 & 4.10 & 5.27 & 3.53 \\
& t5 & \textbf{5.82} & 4.13 & \textbf{5.19} & \textbf{3.50} \\
& t7 & 5.87 & \textbf{4.07} & 5.25 & 3.51 \\
\cline{2-6}
& b1 & 6.01 & 4.19 & 5.45 & 3.62 \\
& b2 & 5.95 & 4.17 & 5.40 & 3.55 \\
& b3 & 5.88 & \textbf{4.13} & 5.34 & 3.51 \\
& b4 & \textbf{5.82} & \textbf{4.13} & \textbf{5.19} & \textbf{3.50} \\
\hline
\multirow{4}{*}{\rotatebox{90}{DexYCB}} 
& c2 & 10.90 & 7.06 & 10.48 & 7.12 \\
& c4 & 9.63 & 6.02 & 9.33 & 6.19 \\
& c6 & 8.41 & 5.16 & 8.17 & 5.61 \\
& c8 & \textbf{8.16} & \textbf{4.81} & \textbf{7.84} & \textbf{5.31} \\
\hline
\end{tabular}
\end{table}
\noindent\textbf{Different Temporal Length.} We examine the performance of UST-Hand with varying temporal lengths in the video sequence. The results on HanCo dataset are shown in~\cref{tab:UST-Hand_performance} rows t1-t7. We find that using 5 frames achieves the best performance. Increasing temporal length from 1 to 5 frames enables the model to capture hand motion patterns and reduce frame-wise jittering. However, extending to 7 frames shows marginal degradation, suggesting that excessively long temporal windows may introduce optimization challenges. We use a temporal length of 5 to balance modeling capacity and computational efficiency.

\noindent\textbf{Different Number of STPT Blocks.} We evaluate the performance of UST-Hand on varying the number of STPT blocks. The results on HanCo dataset are shown in~\cref{tab:UST-Hand_performance} rows b1-b4. The results show that stacking more STPT blocks consistently improves performance. This validates that iterative refinement through multiple blocks enables the model to progressively correct pose errors by repeatedly integrating spatiotemporal information and distributional uncertainty. We use 4 STPT blocks in our final model to achieve optimal performance. 

\noindent\textbf{Different Number of Cameras.} We evaluate UST-Hand across varying numbers of cameras in the multi-view setting. The results on the DexYCB dataset are shown in~\cref{tab:UST-Hand_performance} rows c2-c8, where "c2" indicates the use of two cameras. The results demonstrate that UST-Hand effectively fuses complementary features from diverse viewpoints, boosting overall performance as the number of cameras increases.
\section{More Qualitative Results}
\begin{figure*}
  \centering
  \includegraphics[width=.9\linewidth]{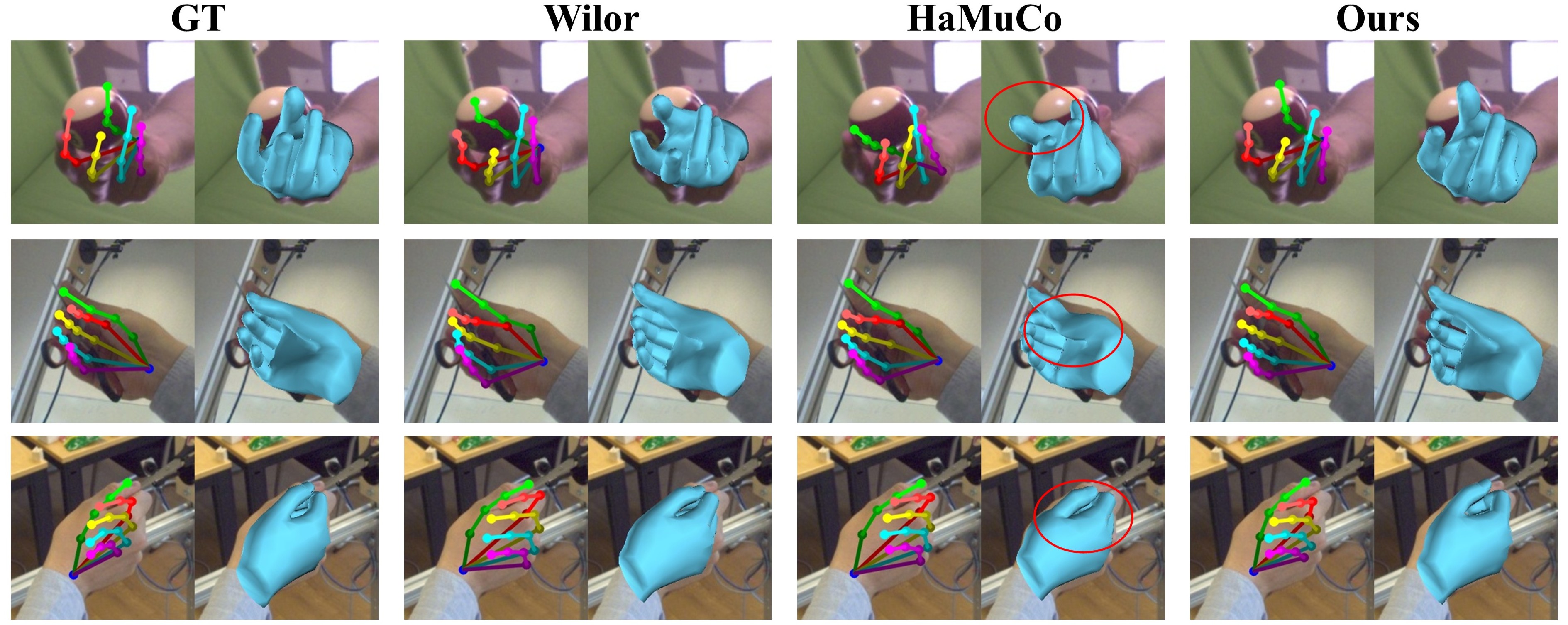}
  \caption{2D joints prediction and 3D mesh prediction between ground-truth, Wilor, HaMuCo, and ours on HanCo dataset.}
  \label{fig:add_hanco}
\end{figure*}

\begin{figure*}
  \centering
  \includegraphics[width=.9\linewidth]{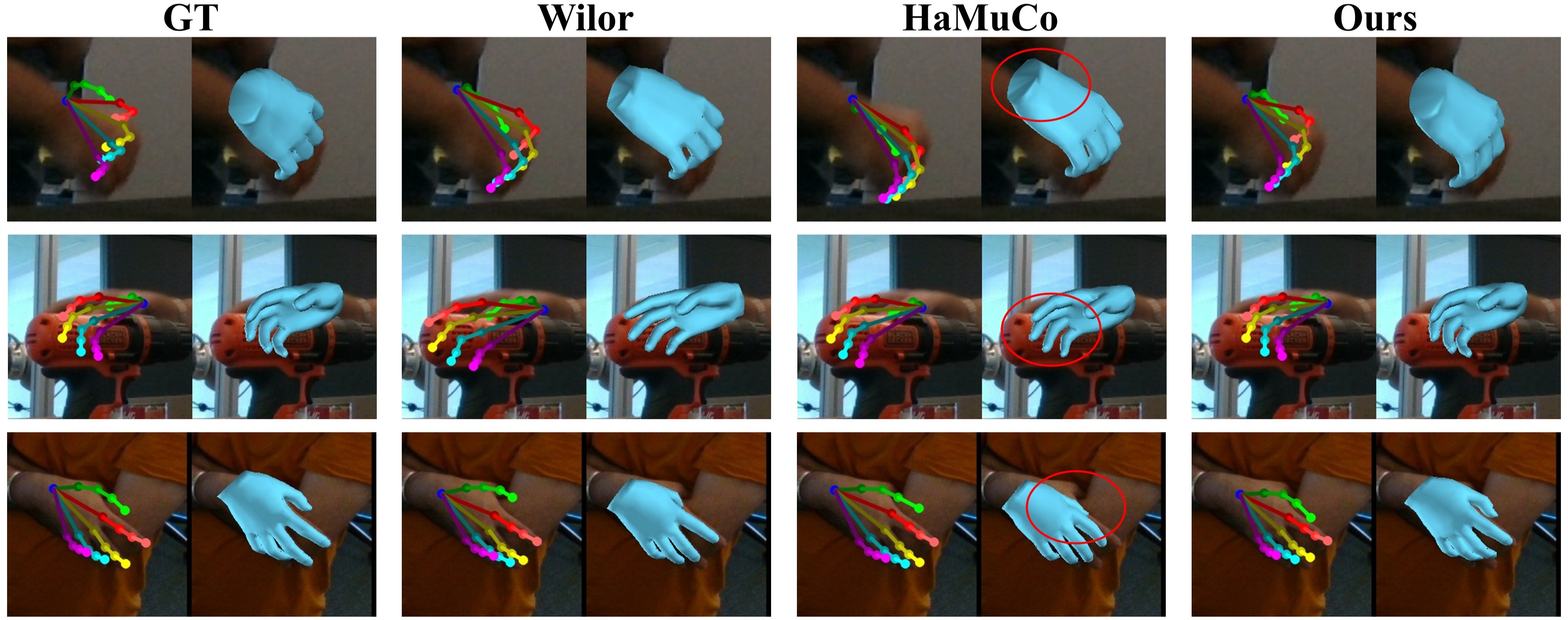}
  \caption{2D joints prediction and 3D mesh prediction between ground-truth, Wilor, HaMuCo, and ours on DexYCB-MV dataset.}
  \label{fig:add_dex}
\end{figure*}

\begin{figure*}
  \centering
  \includegraphics[width=.9\linewidth]{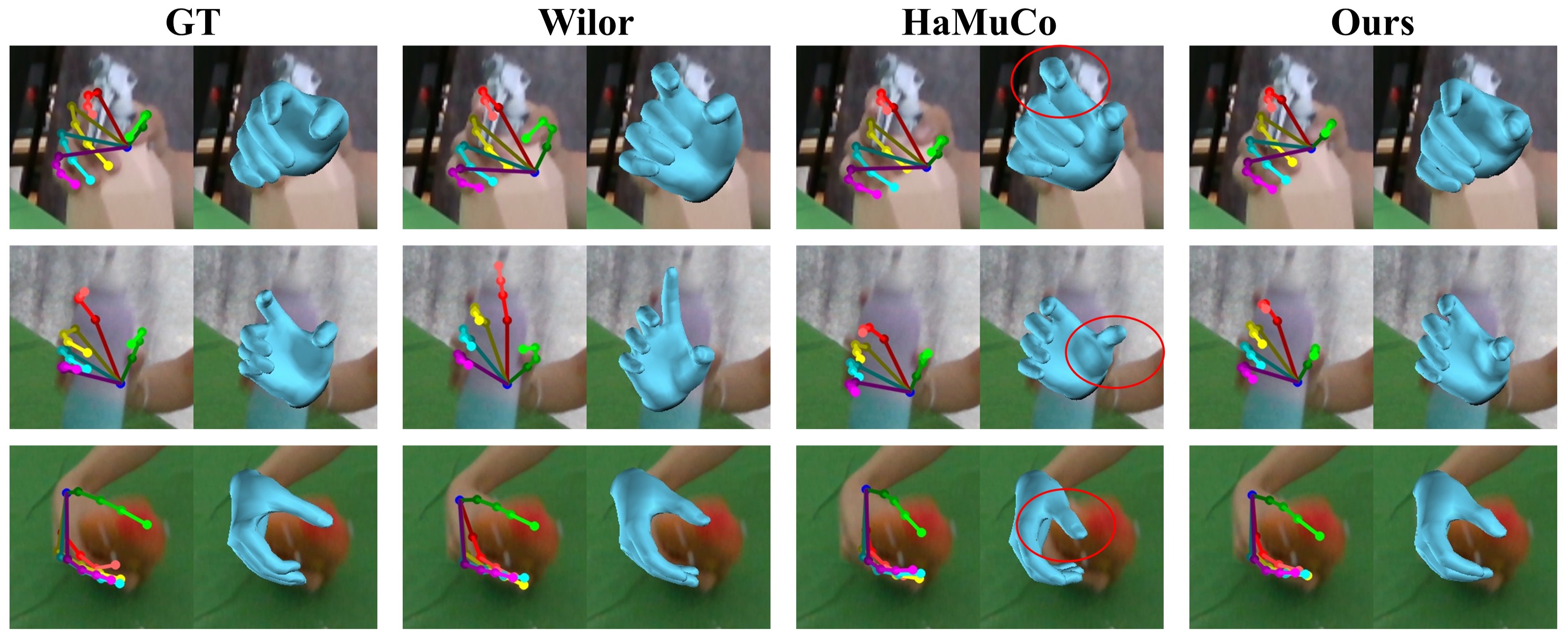}
  \caption{2D joints prediction and 3D mesh prediction between ground-truth, Wilor, HaMuCo, and ours on OakInk-MV dataset.}
  \label{fig:add_oak}
\end{figure*}

\begin{figure*}
  \centering
  \includegraphics[width=1\linewidth]{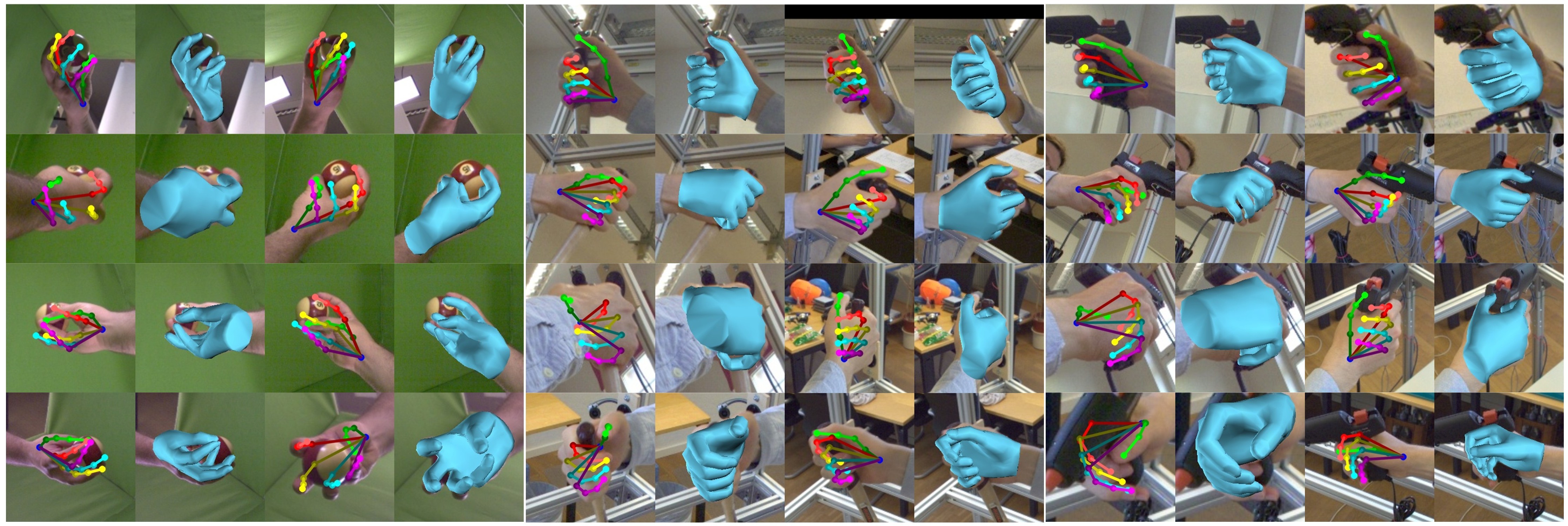}
  \caption{Qualitative results of each view on HanCo dataset.}
  \label{fig:hanco_multi}
\end{figure*}

\begin{figure*}
  \centering
  \includegraphics[width=1\linewidth]{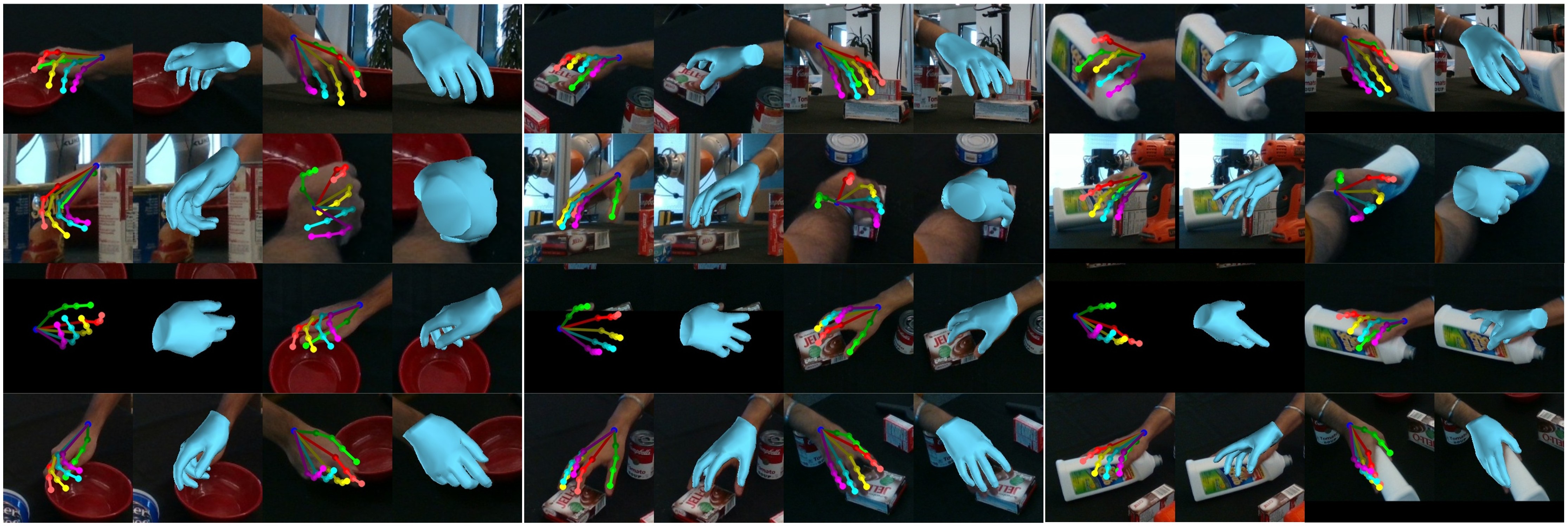}
  \caption{Qualitative results of each view on DexYCB-MV dataset.}
  \label{fig:dex_multi}
\end{figure*}

\begin{figure*}
  \centering
  \includegraphics[width=1\linewidth]{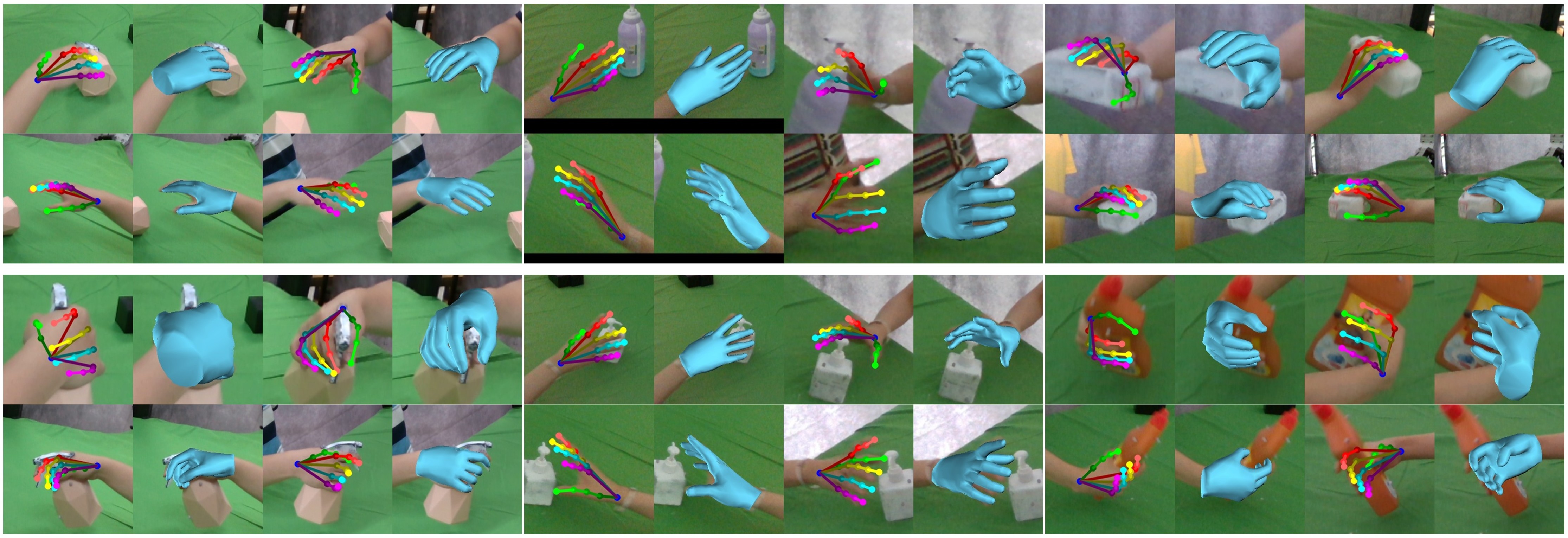}
  \caption{Qualitative results of each view on OakInk-MV dataset.}
  \label{fig:oak_multi}
\end{figure*}

We provide comprehensive qualitative results across all three evaluation datasets to further validate our method's superiority. Specifically, \cref{fig:add_hanco,fig:add_dex,fig:add_oak} compare our method with the SOTA approach HaMuCo, where both models utilize 2D keypoints generated by Wilor for self-supervision. Furthermore, \cref{fig:hanco_multi,fig:dex_multi,fig:oak_multi} explicitly highlight the robustness of UST-Hand in maintaining strict multi-view geometric consistency, even when encountering highly diverse hand poses and complex camera view configurations.



\end{document}